\documentclass[10pt,twocolumn,letterpaper]{article}

\usepackage[pagenumbers]{cvpr} 

\usepackage{graphicx}
\usepackage{amsmath}
\usepackage{amssymb}
\usepackage{booktabs}
\usepackage{color}
\usepackage{multirow}
\usepackage{overpic}
\usepackage{arydshln}
\usepackage{makecell}
\usepackage{times}
\usepackage{epsfig}
\usepackage[T1]{fontenc}
\usepackage{hhline}
\usepackage{xcolor}
\usepackage{pifont}
\usepackage{enumitem}
\usepackage{colortbl}
\usepackage{verbatim}
\usepackage{bm}

\def\ie{\textit{i.e.}}
\def\eg{\textit{e.g.}}

\definecolor{mygray}{gray}{.92}
\definecolor{darkpastelgreen}{rgb}{0.01, 0.75, 0.24}
\definecolor{darkpink}{rgb}{0.91, 0.33, 0.5}

\definecolor{linkcolor}{RGB}{255,0,0}
\definecolor{urlcolor}{RGB}{255,105,180}
\definecolor{citecolor}{RGB}{0, 80, 200}
\definecolor{citecolor1}{RGB}{0,153,255}
\definecolor{teal}{RGB}{49,133,155}
\definecolor{orange}{RGB}{234,112,13}

\usepackage{hyperref}
\hypersetup{colorlinks=true,linkcolor=linkcolor,urlcolor=urlcolor,citecolor=citecolor1}

\newcommand{\cmark}{\ding{51}}%
\newcommand{\xmark}{\ding{55}}%

\usepackage[capitalize]{cleveref}
\crefname{section}{Sec.}{Secs.}
\Crefname{section}{Section}{Sections}
\Crefname{table}{Table}{Tables}
\crefname{table}{Tab.}{Tabs.}

\def\cvprPaperID{2441} 
\def\confName{CVPR}
\def\confYear{2025}
\def\ie{\textit{i.e.}}
\def\eg{\textit{e.g.}}

\begin{document}

\title{Leverage Task Context for Object Affordance Ranking}

\author{Haojie Huang \textsuperscript{1} \quad
Hongchen Luo\textsuperscript{2}\thanks{{Corresponding author. }}  \quad Wei Zhai\textsuperscript{1}\footnotemark[1] \quad Yang Cao\textsuperscript{1,3} \quad Zheng-Jun Zha\textsuperscript{1}
\\
$^{1}$ University of Science and Technology of China \\
$^2$ Northeastern University \\
$^3$ Institute of Artificial Intelligence, Hefei Comprehensive National Science Center  \qquad \\
\small \tt haojie\_h@mail.ustc.edu.cn,luohongchen@ise.neu.edu.cn, \\
\small \tt wzhai056@ustc.edu.cn
}

\maketitle

\begin{abstract}
Intelligent agents accomplish different tasks by utilizing various objects based on their affordance, but how to select appropriate objects according to task context is not well-explored. 
Current studies treat objects within the affordance category as equivalent, ignoring that object affordances vary in priority with different task contexts, hindering accurate decision-making in complex environments.
To enable agents to develop a deeper understanding of the objects required to perform tasks, we propose to leverage task context for object affordance ranking, i.e., given image of a complex scene and the textual description of the affordance and task context, revealing task-object relationships and clarifying the priority rank of detected objects. To this end, we propose a novel Context-embed Group Ranking Framework with task relation mining module and graph group update module to deeply integrate task context and perform global relative relationship transmission. Due to the lack of such data, we construct the first large-scale task-oriented affordance ranking dataset with \textbf{25} common tasks, over \textbf{50k} images and more than \textbf{661k} objects. Experimental results demonstrate the feasibility of the task context based affordance learning paradigm and the superiority of our model over state-of-the-art models in the fields of saliency ranking and multimodal object detection. The source code and dataset will be made available to the public. 


\end{abstract}

\section{Introduction}
\label{sec:intro}

\begin{figure}[t]
	\centering
		\begin{overpic}[width=0.99\linewidth]{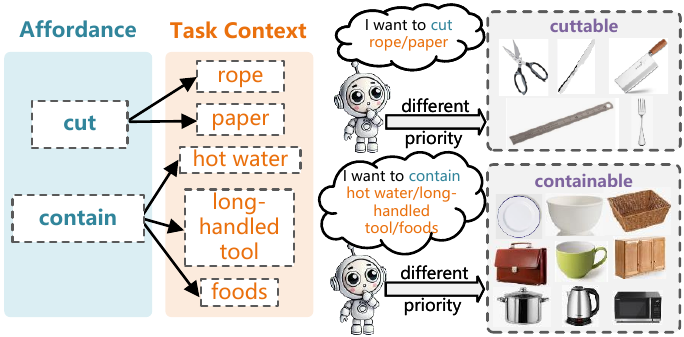}
	\end{overpic}
        \captionsetup{aboveskip=0pt}
        \captionsetup{belowskip=0pt}
	\caption{Task Context represents the subsequent purpose of a task. For the same affordance, different task contexts lead to varying priority of objects.}
	\label{figure1}
\end{figure}

Affordance learning has become a well-researched field in recent years\cite{pieropan2015functional,koppula2014physically,ugur2011going,ye2017can}, mainly focusing on tagging affordance labels to objects\cite{sun2010learning,varadarajan2012afrob} or identifying their interactive areas\cite{nagarajan2019grounded,do2018affordancenet,sawatzky2017adaptive}, but does not consider how an object’s affordance varies across different task contexts and the priority of different objects within the same task context.
Here, the task context denotes the specific requirements for using the object, as shown in Fig. \ref{figure1}. Intelligent agents are context-sensitive systems that adapt more effectively to real-world environments through contextual learning, which is advantageous for scene understanding\cite{zhu2015understanding,chuang2018learning}, VR/AR\cite{steffen2019framework,zheng2018affordances,shin2022does} and embodied AI\cite{ahn2022can,ge2024behavior}.


\begin{figure*}[t]
	\centering
		\begin{overpic}[width=1.0\linewidth]{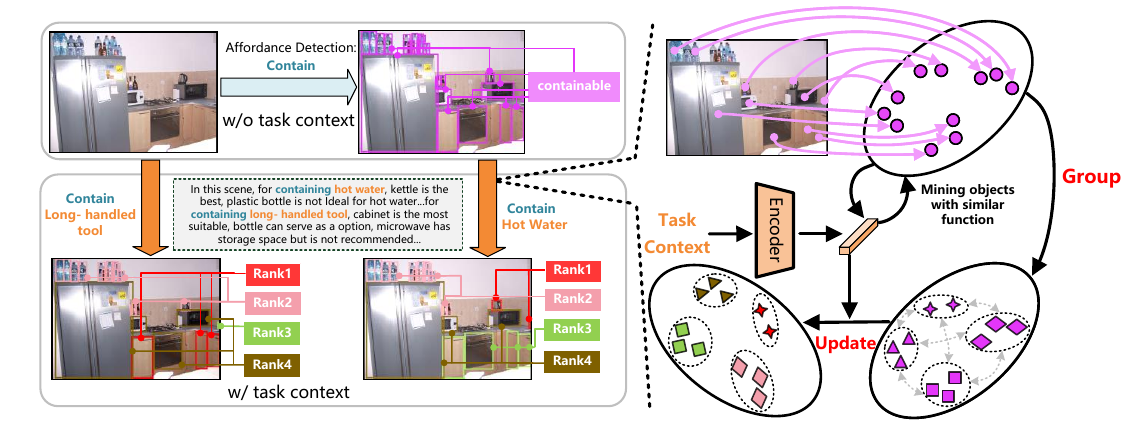}
		\put(29.6,-0.8){\textbf{(a)}}
		\put(73,-0.8){\textbf{(b)}}
	\end{overpic}
      \captionsetup{aboveskip=6pt}
     \captionsetup{belowskip=0pt}
	\caption{\textbf{Motivation.}
	 (a) This paper considers the impact of different task contexts on object affordances and achieves a more fine-grained understanding of objects through ranking. (b) We deeply fuse task context to group candidate objects and perform context conditioned graph update to obtain the correct relative ranking. 
	}
	\label{motivation}
\end{figure*}

Some previous works explored task-driven affordance detection to select task-relevant objects\cite{li2022toist,chen2024taskclip,chen2024vltp,qu2024rio,sawatzky2019object}. However, these works primarily treat all objects equally, without modeling the priority relationships between them in relation to the task context and do not account for how the same object varies across different task contexts, which makes it difficult for the agent to make the correct choices in the environment and hinders the agent's robustness and flexibility. Human can model the prioritization of information based on task requirements to better utilize cognitive resources \cite{kahneman1973attention}. To endow the agents with this humanlike common sense, we consider leveraging task context for object affordance ranking in this paper(Fig. \ref{motivation} (a)), \ie, given image of a complex scene and the textual description of the affordance and task context, revealing task-object relationships and clarifying the priority rank of detected objects. 

This challenging task includes several issues that should be properly addressed. \textbf{1) Affordance-context adaptation}. Our ranking requires identifying the affordance of objects, which is a dynamic property that varies with different task contexts, also leading to changes in priority ranks. 
The existing fixed single-mapping approach may lead to misinterpretation of objects within the scene, making it difficult to accurately select appropriate tools, we need to deeply integrate task features to dynamically assess the affordance of objects. 
\textbf{2) Task-objects heterogeneity}. Due to the hierarchical structure of real-world scene, task-related objects are not confined to a single instance or category but are instead organized into functional groups \cite{huang2018holistic, zhao2013scene, liu2017single}. Here, as a more detailed concept than previous ones, a functional group refers to a collection of objects that offer the same affordance and share an identical priority rank based on the task context. Real-world environments are highly diverse, with objects arranged in numerous ways and the number of instances and categories within the group may vary. This heterogeneity makes it difficult for algorithms to identify consistent functional groups across different scenes, not to mention modeling the priority rank among these groups.


To overcome these problems, we propose the \textbf{C}ontext-embed \textbf{G}roup \textbf{R}anking (\textbf{CGR}) framework to extract task-specific visual features and modeling the priority relationships between them. It utilizes an end-to-end encoder-decoder architecture to simultaneously detect task-related objects and assign their ranks using graph-based relative relationship computation, as shown in Fig. \ref{motivation} (b). Firstly, a \textbf{T}ask \textbf{R}elation \textbf{M}ining (\textbf{TRM}) module is introduced to deeply integrate affordance and task context with image features by bidirectional feature fusion and affordance-guided query selection, which can minimize computations on irrelevant regions and focus on the particular affordance. Then, We design a novel \textbf{G}raph \textbf{G}roup \textbf{U}pdate (\textbf{GGU}) strategy that incorporates group information aggregation and context conditioned graph update to organize the objects into several functional groups and calculate their relative rankings. In addition, a new grouping ranking loss is proposed to better train our group update branch by penalizing incorrect groupings. 



To accomplish the objectives presented in this paper, we propose the \textbf{T}ask-oriented \textbf{A}ffordance \textbf{R}anking (\textbf{TAR}) dataset, consisting of 50,404 real-world images from 25 tasks and 70 object categories with 11,651 possible ranking scenarios(unique category combinations represents ranking scenarios). The TAR is compared with other related datasets in Table \ref{Table:relevant datasets}. Specifically, ranking is meaningful only in complex scene. Our TAR dataset contains 81,643 affordance-objects pairs, totaling 661k object instances, with an average of over 8 objects to be ranked per pair. We establish six baselines in related fields on the TAR dataset and results demonstrate the superiority of our proposed method in understanding object affordance and ranking them according to task context. 



\textbf{Contributions:} (1) We introduce a new affordance learning paradigm: leverage task context for object affordance ranking, which is designed to enable intelligent agents to perform tasks more accurately in real-world environments. (2) A novel context-embed group ranking framework is proposed, which mitigates the dynamic properties of object affordance and groups objects with similar function priorities, akin to human cognitive processes, while performing priority ranking by incorporating task context. (3) We collect and annotate a large scale dataset, TAR, to support the proposed task, which includes a variety of complex scenes and their corresponding rankings. Experimental results show that our model can achieve the best performance among other state-of-the-art models and can serve as a robust baseline for future research.


\section{Related Works}
\label{sec:relation work}
\subsection{Affordance Learning}
Affordance learning is a key component of intelligent systems, as it endows agents with the crucial ability to understand and adapt to their environment, which is essential for human-object interaction and agent decision-making. Early works\cite{li2023locate,myers2015affordance,nguyen2017object,roy2016multi} focused on learning an intrinsic object-affordance mapping, assuming a one-to-one correspondence. However, these approaches struggle to handle situations in real-world scenarios where objects may possess multiple affordances. Researchers have tried various methods to tackle this practical issue. Some use reinforcement learning to simulate the way humans learn affordance through practice and feedback\cite{yang2023recent,nagarajan2020learning}, some learn affordance through human demonstrations\cite{bahl2023affordances,liu2022joint,chen2023affordance} and some leverage human-object interaction to localize object affordance\cite{luo2023leverage,luo2022learning,yang2024egochoir}. However, these studies are limited to the visual features of the objects themselves, overlooking the fact that different tasks can lead to varying affordances for the same objects. 
Then, some task-driven benchmark were constructed, and subsequent works focused on task-driven affordance detection\cite{zhu2015understanding,tang2023cotdet,wolfel2018grounding, lu2022phrase}. Nevertheless, these studies did not take into account either the impact of varying task contexts on object affordances or the priority difference between candidate objects. To this end, this paper considers ranking the affordances of different objects based on tasks to help agents make better decisions.

\subsection{Object Ranking}
Recently, saliency ranking\cite{islam2018revisiting,siris2020inferring,yildirim2020evaluating,lin2022rethinking} has emerged as a challenging task in computer vision aiming not only to detect salient objects within images but also to rank them based on the degree of visual stimulation they provide to viewers. Similarly, camouflage ranking\cite{lv2021simultaneously,lamdouar2023making} aims to rank the level of camouflage by how they blend into their background. The ranking results in the tasks mentioned above are mainly determined by factors such as object size, color, and contrast with the overall environment, without integrating any semantic context. In contrast, our task is more challenging, as it determines object priority not only by assessing the visual features of the objects but also by considering how well their affordances align with task requirements. Additionally, their ranking follows a strict sequence order. However, in our task, the objects capable of completing the task typically appear in groups, and the detected objects, groupings, and rankings in the same image can vary depending on the specific task, making our task even more complex. 



\begin{table}[!t]
    \centering
  \scriptsize
  \renewcommand{\arraystretch}{1.}
  \renewcommand{\tabcolsep}{5.4pt}
    \captionsetup{aboveskip=0pt}
  \captionsetup{belowskip=0pt}
   \caption{\textbf{Comparison with previous affordance learning datasets and task-driven affordance detection datasets. } Obj: total number of object instances. Aff: number of affordance classes. Rank: whether there are rank annotations}
\label{Table:relevant datasets}
  \begin{tabular}{c|ccc|cc|c}
\hline
\Xhline{2.\arrayrulewidth}
\textbf{Dataset} & \textbf{Images}  & \textbf{Classes} & \textbf{Obj} &  \textbf{Aff} & \textbf{Tasks} & \textbf{Rank}   \\
\hline
\Xhline{2.\arrayrulewidth}
 ADE-Aff \cite{chuang2018learning}  & 10,000  & 150  & 22k & 7  & 7  & \xmark \\
 COCO-Task \cite{sawatzky2019object} & 39,724 & 49 & 65k & 12 & 14 & \xmark  \\
 PAD \cite{luo2021one} & 4,002 & 72 & 4k & 31 & \xmark & \xmark  \\
 PADv2 \cite{zhai2022one} & 30,000 & 103 & 30k & 39 & \xmark & \xmark  \\
 AGD20k \cite{luo2022learning} & 23,816 & 50 & 23k & 36 & \xmark & \xmark \\
 RIO \cite{qu2024rio} & 40,214 & 69 & 130k & >100 & >100 & \xmark  \\
\hline
\rowcolor{mygray}
TAR (Ours) & 50,404 & 70 & 661k & 11 & 25 & \cmark \\
\hline
\Xhline{2.\arrayrulewidth}
    \end{tabular}
\end{table}

\begin{figure*}[t]
	\centering
		\begin{overpic}[width=0.98\linewidth]{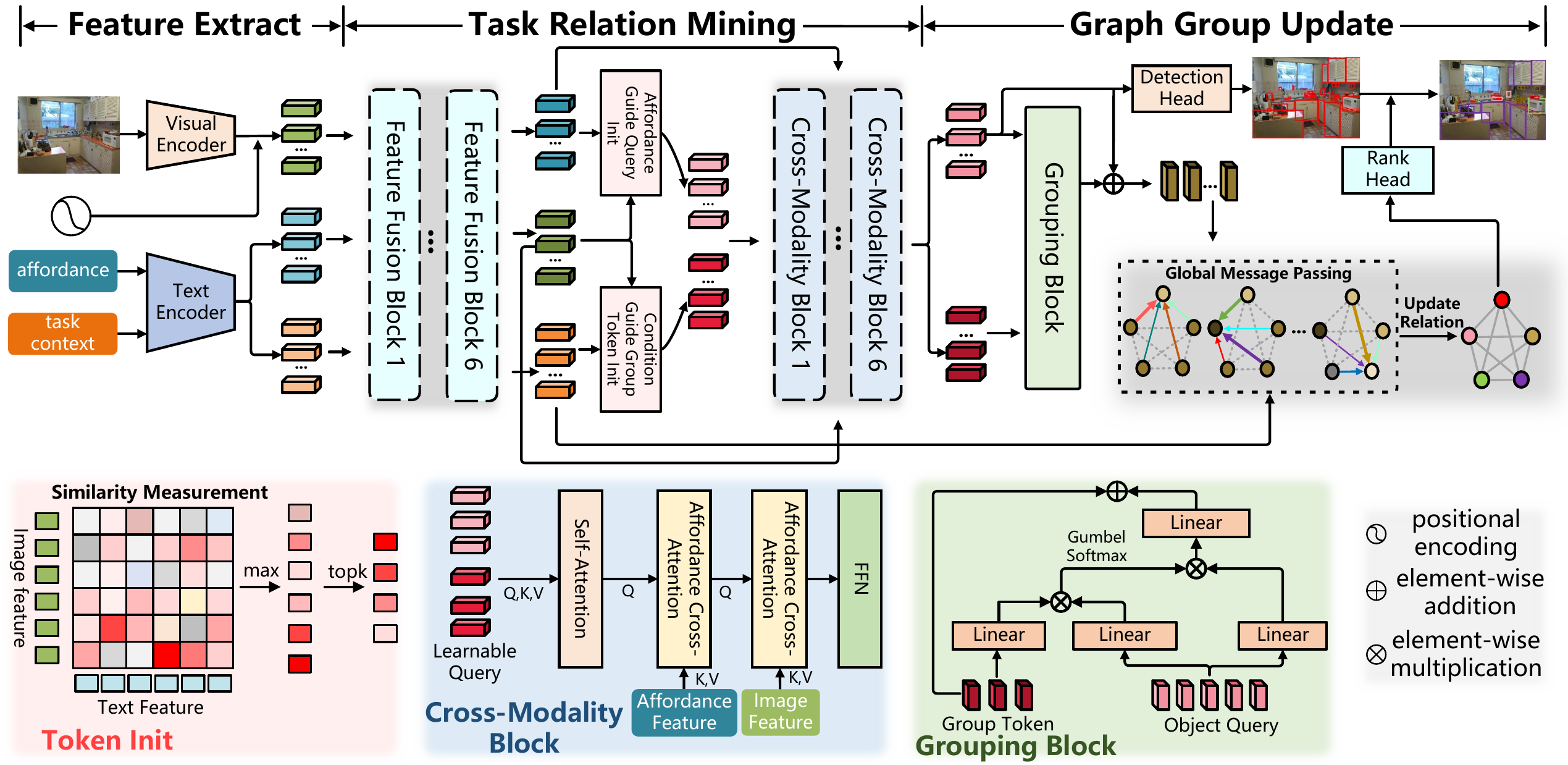}
        \put(18.3,43.0){\scriptsize{$\bm{F_I}$}}
        \put(18.3,36.1){\scriptsize{$\bm{F_a}$}}
        \put(18.3,29.0){\scriptsize{$\bm{F_c}$}}
        \put(36.2,43.5){\scriptsize{$\bm{T_a}$}}
        \put(36.2,36.3){\scriptsize{$\bm{I}$}}
        \put(36.2,29.0){\scriptsize{$\bm{T_c}$}}
        \put(44.7,39.3){\scriptsize{$\bm{O}$}}
        \put(44.5,26.6){\scriptsize{$\bm{G}$}}
        \put(61.0,42.7){\scriptsize{$\bm{\tilde{O}}$}}
        \put(60.8,22.8){\scriptsize{$\bm{\tilde{G}}$}}
        \put(76.0,39.0){\scriptsize{$\bm{\Theta}$}}
        \put(64.8,6.1){\scriptsize{$\bm{W_q}$}}
        \put(73.6,9.5){\scriptsize{$\bm{W_k}$}}
        \put(82.3,9.5){\scriptsize{$\bm{W_v}$}}
        \put(78.3,16.5){\scriptsize{$\bm{W_o}$}}
	\end{overpic}
            \captionsetup{aboveskip=2pt}
        \captionsetup{belowskip=0pt}
	\caption{\textbf{Overview of Context-embed Group Ranking (CGR) Model.}
	It first extracts image features $F_I$ and task features $F_a,F_c$ separately, then aligns them through the TRM module (Sec. \ref{TRM}) and get gathered object queries $\tilde{O}$ and Group Tokens $\tilde{G}$. Subsequently, the GGU (Sec. \ref{GGU}) groups $\tilde{O}$ using $\tilde{G}$ and aggregates group information, then performs global message passing with context features $T_c$ to determine final ranking results.
	}
	\label{model}
\end{figure*}
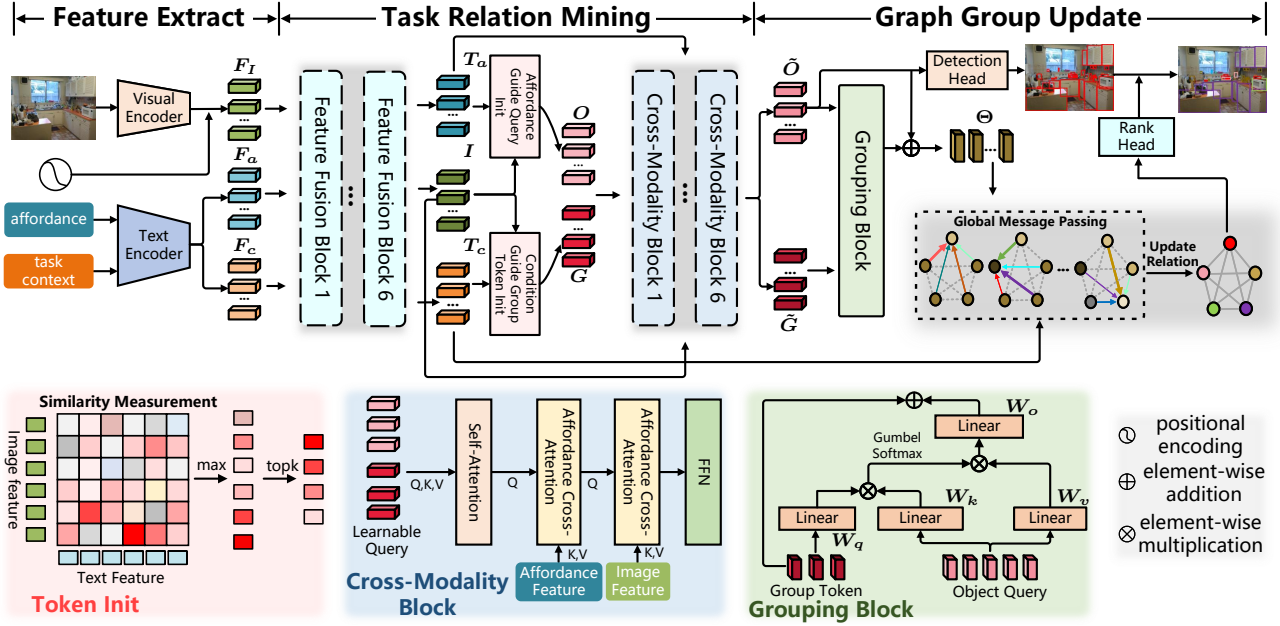

\section{Method}
\label{sec:method}
\subsection{Overview}
Our CGR (Fig. \ref{model}) is designed to identify all task-related objects and determine their hierarchical priority ranks in an end-to-end fashion. In detail, given a sample $\{\mathcal{S}, I\}$, $\mathcal{S}$ represents the textual task input which consisting of an affordance part $S_a$ and a context part $S_c$, $I \in R^{3 \times h \times w}$ is an RGB image, CGR will predict the bounding boxes of $N$ task-related objects in $I$ and further predict their ranks. Initially, CGR employs a pretrained Swin Transformer \cite{liu2021swin} to extract vanilla image features, and then added with 2D positional encodings to conserve the spatial information, yielding the image features $F_I$. For the text input, a shared Bert model \cite{devlin2018bert} is used to extract affordance features $F_a$ and context features $F_c$ respectively. Then, $F_I$, $F_a$ and $F_c$ will be fed into the TRM module (see in Sec. \ref{TRM}) for integration of task context and aggregation of group tokens by bidirectional cross-attention and multimodal group decoder, obtaining object features $\tilde{O}$ and group features $\tilde{G}$. The object features $\tilde{O}$ are then passed to a detection head to get the bounding box of candidate objects. Next, we adopt a Graph Group Update strategy (see in Sec. \ref{GGU}) to group objects and disseminate competition relationships among different groups. Finally, we predict the rank scores by a simple rank head and assign them to corresponding bounding boxes.




\subsection{Task Relation Mining}
\label{TRM}
Our TRM module achieves deep interaction of task information through a two-step feature aggregation process. Firstly, we map $F_I$ and $F_a$, $F_c$ to the same dimension using a convolution layer and a linear layer respectively. Then, a deformable self-attention mechanism is used to enhance flattened image features and a classic self-attention mechanism is used to enhance text features, obtaining $\Bar{F_I}$, $\Bar{F_a}, \Bar{F_c}$. 
Given that the features originate from different modalities and are encoded by distinct encoders, we concatenate $\Bar{F_a}, \Bar{F_c}$ to get the whole textual task features $\Bar{F_T}$ and sequentially utilize text-to-image and image-to-text cross-attention layers to align visual and textual features: 
\begin{equation}
\small
    F_{iq}=F_IW^1, F_{iv}=F_IW^2, F_{tq}=F_TW^3, F_{tv}=F_TW^4,
\label{equ2}
\end{equation}
\begin{equation}
\small
    I=SoftMax(F_{iq}F_{tq}^T/\sqrt{d})F_{tv},
\label{equ3}
\end{equation}
\begin{equation}
\small
    T_a, T_c=SoftMax(F_{tq}F_{iq}^T/\sqrt{d})F_{iv},
\label{equ4}
\end{equation}
where $W^{1\sim4}$ are projection weights, and in $F_{\{i,t\}, \{q,v\}}$ the first subscript denotes image or text, and the second specifies query or value in the attention mechanism.

Secondly, similar to many DETR-like models\cite{carion2020end}, we use object queries to aggregate the image features $I$ obtained from the encoder, which are ultimately used to predict object bounding boxes. However, unlike the commonly used randomly initialized query embeddings, we utilize fused affordance features $T_a$ to select the most relevant visual features as queries $O$, allowing the task input to guide the generation of candidate objects. Meanwhile, we use a similar approach by leveraging $T_c$ to initialize group tokens \cite{xu2022groupvit} $G$ to embed context information for better grouping:
\begin{equation}
\small
    O=I[topK(Max(IT_a^T))], G=I[topK(Max(IT_c^T))],
\label{equ5}
\end{equation}
where $[\cdot]$ indicates indexing operation. 
Following, $O$ and $G$ are fed into several cross-modality attention blocks for the second step of feature aggregation\cite{liu2023grounding,ren2024grounding}. Initially, we use a self-attention layer $f_{og}$ to enable $G$ to model global information from $O$, facilitating subsequent object grouping:
\begin{equation}
\small
    [\Bar{O},\Bar{G}]=f_{og}([O,G]),
\label{equ6}
\end{equation}
where $[\cdot]$ indicates concatenation. Next, $\Bar{O}$ and $T_a$ are fed into a cross-attention layer to further integrate affordance information and then $\Bar{O}$ and $\Bar{G}$ are concatenated and passed into another cross-attention layer with $I$ to combine image features jointly, obtaining gathered object queries $\tilde{O}$ and group tokens $\tilde{G}$. 

\begin{figure*}[t]
	\centering
		\begin{overpic}[width=0.98\linewidth]{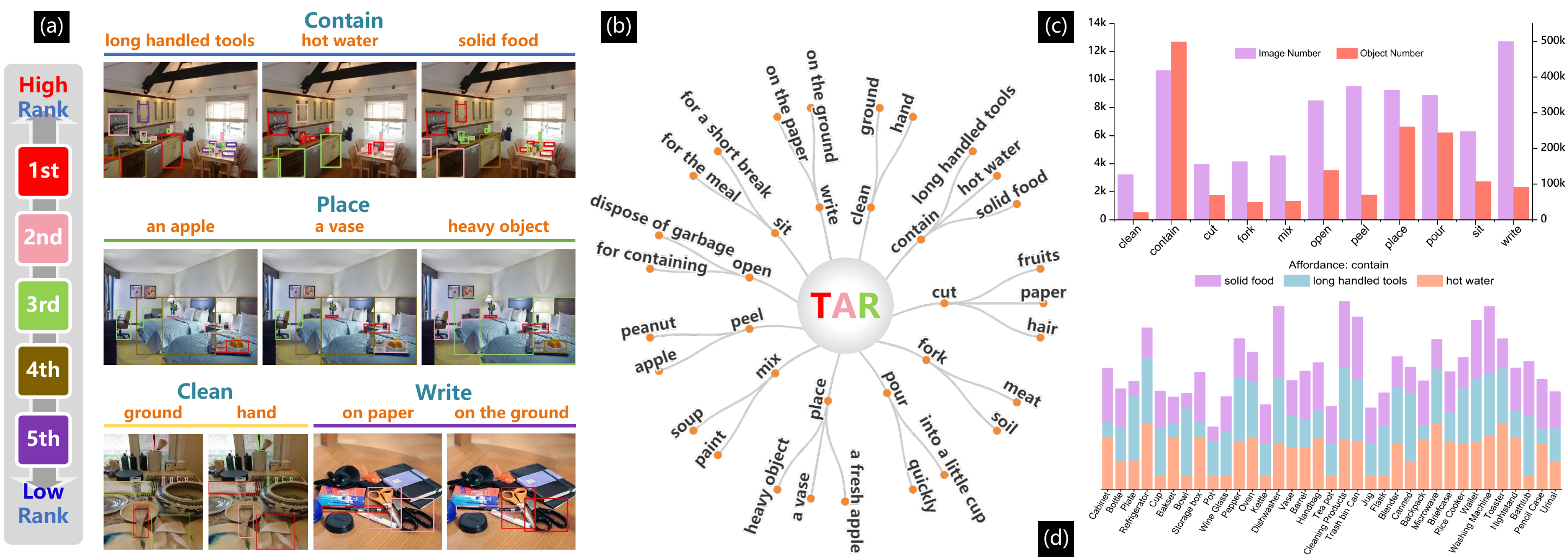}
	\end{overpic}
         \captionsetup{aboveskip=1pt}
         \setlength{\belowcaptionskip}{0pt} 
	\caption{\textbf{The properties of the TAR dataset.} 
 (a) Ranking annotation samples of affordance ``contain'' and ``place'' with different task contexts (b) Total affordance verbs and corresponding task contexts (c) Distribution of image and object numbers across different affordance (d) Object average rank across different task contexts within affordance ``contain''.
	}
	\label{data}
\end{figure*}

\subsection{Graph Group Update}
\label{GGU}
A simple method for predicting affordance ranking is to directly input $\tilde{O}$ into a linear layer to regress the ranking logits. However, this approach fails to capture the global relative priority relationships, and the absence of a pre-grouping operation may lead to incorrect competition among objects with the same rank. To this end, GGU first inputs $\tilde{O}$ and $\tilde{G}$ into the group block to divide the detected objects into several groups using Gumbel-Softmax\cite{jang2016categorical}:
\begin{equation}
\small
    g_{i,j}=\frac{exp((\tilde{O_i}W^o(\tilde{G_j}W^g)^T+\pi_j) / \tau)}{\sum_k^Nexp((\tilde{O_i}W^o(\tilde{G_k}W^g)^T+\pi_k) / \tau)},
\label{equ7}
\end{equation}
where $g_{i,j}$ represents the probability that the $i$-th object belongs to the $j$-th group, $\pi_1...\pi_k$ are i.i.d samples drawn from the Gumbel(0,1) distribution. Then, we use argmax to obtain the specific grouping result $\hat{g}$, and apply the \textit{hard assignment} strategy proposed in \cite{xu2022groupvit} to make this discrete process differentiable. After grouping, we average the features of all objects within the same group to obtain the group embedding $\boldsymbol{\Phi}$, which is then added to the corresponding object features to get the grouped object features \boldsymbol{$\Theta$}, ensuring that objects within the same group share the same grouping characteristics:
\begin{equation}
\small
    \boldsymbol{\Phi_j}=\frac{1}{K}\sum_{k\in D}\tilde{O_k},D=\{x|\hat{g}[x]=j\},K=|D|,
\label{equ8}
\end{equation}
\begin{equation}
\small
\boldsymbol{\Theta_i}=\tilde{O_i}+\boldsymbol{\Phi}[\hat{g}[i]].
\label{equ9}
\end{equation}
We establish a fully connected graph $\mathcal{G}=(\mathcal{V},\mathcal{E})$ to perform relational operations, where $\mathcal{V}=\{\boldsymbol{\Theta_1},...,\boldsymbol{\Theta_N}\}$, $\mathcal{E}$ is a matrix of ones, representing that there is an edge between any two instances. During the forward computation, for entity $\boldsymbol{\Theta_i}$, we first calculate its connection weights $w_{ji}$ with all entities $\boldsymbol{\Theta_j}$ ($j\in1...N$) predicated on context features $T_c$:
\begin{equation}
\small
    w_{ji}=\underset{j}{Softmax}(((\boldsymbol{\Theta_j}W^5)\odot(T_cW^6))(\boldsymbol{\Theta_i}W^7)^T).
\label{equ10}
\end{equation}
Then each entity $\boldsymbol{\Theta_j}$ passes a message $m_{ji}$ to entity $\boldsymbol{\Theta_i}$ and $\boldsymbol{\Theta_i}$ sum up all the messages and update the relational information:
\begin{equation}
\small
    m_{ji}= w_{ji} \cdot \boldsymbol{\Theta_j}W^8,
\label{equ11}
\end{equation}
\begin{equation}
\small
    \boldsymbol{\Theta_i^r}= \boldsymbol{\Theta_i} + (\sum_{j=1}^Nm_{ji})W^9.
\label{equ12}
\end{equation}
Finally, a simple ranking head is adopted to regress the rank scores by $\boldsymbol{\Theta^r}$, which are assigned to corresponding bounding boxes, obtaining the results as shown in Fig. \ref{model}.

\subsection{Loss Function}
Total loss consists of the detection loss and the ranking loss. The detection loss comprises classification loss $\mathcal{L}_{ce}$ and bounding box loss $\mathcal{L}_{box}$. For $\mathcal{L}_{ce}$, we utilize focal loss\cite{ross2017focal} to mitigate the imbalance of positive and negative samples caused by the small proportion of task-relevant objects among all candidate queries. $\mathcal{L}_{box}$ incorporates the commonly used $\ell_1$ loss and generalized IoU loss\cite{rezatofighi2019generalized}. For the ranking loss $\mathcal{L}_{rank}$, despite some previous studies\cite{liu2021instance,qiao2024hypersor} used pair-wise ranking loss to address dynamic ranking levels, but the loss did not effectively penalize cases of incorrect grouping. Thus, we improve the previous ranking loss to better train our grouping mechanism by introducing group penalty parameter $\rho$. Specifically, we continue to use the pair-wise calculation mechanism, and for an image containing N instances, the predicted rank scores are $\{r_1^p, r_2^p...r_N^p\}$, the gt ranks are $\{r_1^{gt}, r_2^{gt}...r_N^{gt}\}$, we extract all $C_N^2$ pairwise combinations to calculate the loss as:
\begin{equation}
\small
    d_p=(-r_i^p+r_j^p), d_{gt}=r_i^{gt}-r_j^{gt},
\label{equ13}
\end{equation}
\begin{equation}
\small
    d_{ij}=d_p \cdot sign(d_{gt}) + (1-sign(d_{gt})) \cdot abs(d_p),
\label{equ14}
\end{equation}
\begin{equation}
\small
\mathcal{L}_{rank}=\sum_{ij}^{C_N^2}\beta_{ij}log(1+exp(d_{ij})),
\label{equ15}
\end{equation}
\begin{equation}
\small
    \beta_{ij}=
    \begin{cases}
    \rho & r_i^{gt}=r_j^{gt} \\
    \frac{r_i^{gt}-r_j^{gt}}{\sum_{ij}^{C_N^2}r_i^{gt}-r_j^{gt}} & r_i^{gt} \neq r_j^{gt} .
    \end{cases}
\label{equ16}
\end{equation}
Eventually, the total loss is formulated as:
\begin{equation}
\small
\mathcal{L}=\lambda_1\mathcal{L}_{ce}+\lambda_2\mathcal{L}_{l1}+\lambda_3\mathcal{L}_{giou}+\lambda_4\mathcal{L}_{rank},
\label{equ17}
\end{equation}
where $\lambda_1$, $\lambda_2$, $\lambda_3$ and $\lambda_4$ are hyper-parameters to balance the total loss.

\section{Proposed Dataset}
\subsection{Dataset Collection}
Our dataset is primarily obtained from Objects365\cite{shao2019objects365}, which contains a vast number of images and categories. We first define 11 commonly used affordance verbs in real-world scenarios, each associated with at least two task contexts, and combine them into 25 unique tasks in verb-object or verb-adverb structures, as illustrated in the Fig. \ref{data} (b). Subsequently, we select object categories relevant to each affordance verb, \eg, ``cup'', ``cabinet'', ``refrigerator'' etc are associated to ``contain'', and choose images containing at least two alternative categories for ranking. Since different categories may belong to the same functional group and share the same rank, we exclude instances that have only one ranking level. In total, we collect 50,404 images and 81,643 affordance-objects pairs.

\subsection{Dataset Annotation}
For each image, we first identify the tasks it may correspond to based on the objects present in the image. Then, we assess whether the features of the object instances in the image, such as size, shape, material, etc., align with the task context, and assign ranking levels based on their relevance, with smaller numbers representing higher priority. 
In this case, varying contexts in the same affordance correspond to different ranks, which means each image can have multiple rank annotations as shown in Fig. \ref{data} (a), effectively increasing the number of training instances and also consistent with real-world scenarios. Furthermore, given that scenes inevitably contain objects completely unrelated to the task, we assign these instances a very low ranking level to mark them as irrelevant, allowing them to be filtered out during the inference phase based on a defined threshold. Ultimately, the dataset comprises an impressive total of 187,129 instances, with 3,372,893 objects to be ranked. 

\subsection{Statistic Analysis}
The distribution of image and object number for each affordance is shown in the Fig. \ref{data} (c), demonstrating the abundant training instances available for each task in our dataset. To more clearly illustrate how object priority levels vary with task contexts, we calculate the average rank of each object category under the three contexts for the affordance ``contain'' as shown in the Fig. \ref{data} (d). It is evident that the average rank of objects significantly differ across the different contexts, posing a considerable challenge for our ranking task.

\begin{table*}[!t]
    \centering
  \footnotesize
  \renewcommand{\arraystretch}{1.}
  \renewcommand{\tabcolsep}{11.6pt}
  \captionsetup{aboveskip=1pt}
  \captionsetup{belowskip=1pt}
   \caption{\textbf{The results of different methods on TAR.} The best results in the table are highlighted in \textbf{bold}. The $\textcolor{darkpink}{\diamond}$ defines the relative improvement of our method over other methods. ``\textcolor[rgb]{0.6,0.1,0.1}{Dark red}'' and ``\textcolor[rgb]{0.99,0.5,0.0}{Orange}'' represent saliency ranking and multimodal object detection models. Task-acc and Task-recall are shown in percentage.}
   \label{ce_res}
  \begin{tabular}{r||cccccc}
    \hline
    \Xhline{2.\arrayrulewidth}
   \textbf{Methods} & $\text{SSOR}$ & $\text{SA-SOR} $ & $\text{mAP@50} $  & $\text{ARI}$ & $\text{Task-acc}$ & $\text{Task-recall}$
\\   \hline
\Xhline{2.\arrayrulewidth}
  \textcolor[rgb]{0.6,0.1,0.1}{SOR}  \cite{fang2021salient} & 
  $0.497\textcolor{darkpink}{\scriptstyle~\diamond34.2\%}$ & $0.321\textcolor{darkpink}{\scriptstyle~\diamond71.3\%}$ & $24.6\textcolor{darkpink}{\scriptstyle~\diamond63.4\%}$ & $0.359\textcolor{darkpink}{\scriptstyle~\diamond112.5\%}$ & $84.7\textcolor{darkpink}{\scriptstyle~\diamond10.7\%}$ & $78.1\textcolor{darkpink}{\scriptstyle~\diamond16.0\%}$ \\
  \textcolor[rgb]{0.6,0.1,0.1}{IRSR} \cite{liu2021instance} & 
  $0.526\textcolor{darkpink}{\scriptstyle~\diamond26.8\%}$ & $0.414\textcolor{darkpink}{\scriptstyle~\diamond32.9\%}$ & $25.0\textcolor{darkpink}{\scriptstyle~\diamond60.8\%}$ & $0.452\textcolor{darkpink}{\scriptstyle~\diamond68.8\%}$ & $83.2\textcolor{darkpink}{\scriptstyle~\diamond12.7\%}$ & $74.1\textcolor{darkpink}{\scriptstyle~\diamond22.2\%}$\\
  \textcolor[rgb]{0.6,0.1,0.1}{QAGNet} \cite{deng2024advancing} & $0.509\textcolor{darkpink}{\scriptstyle~\diamond31.0\%}$ & $0.406\textcolor{darkpink}{\scriptstyle~\diamond35.5\%}$ & $23.1\textcolor{darkpink}{\scriptstyle~\diamond74.0\%}$ & $0.414\textcolor{darkpink}{\scriptstyle~\diamond84.3\%}$ & $90.1\textcolor{darkpink}{\scriptstyle~\diamond4.1\%}$ & $82.3\textcolor{darkpink}{\scriptstyle~\diamond10.0\%}$ \\
  \hline   
  \textcolor[rgb]{0.99,0.5,0.0}{MDetr} \cite{kamath2021mdetr} & $0.483\textcolor{darkpink}{\scriptstyle~\diamond38.1\%}$ & $0.509\textcolor{darkpink}{\scriptstyle~\diamond8.1\%}$ &  $29.6\textcolor{darkpink}{\scriptstyle~\diamond35.8\%}$ & $0.394\textcolor{darkpink}{\scriptstyle~\diamond93.7\%}$ & $83.3\textcolor{darkpink}{\scriptstyle~\diamond12.6\%}$ & $70.7\textcolor{darkpink}{\scriptstyle~\diamond28.1\%}$ \\
   \textcolor[rgb]{0.99,0.5,0.0}{OVDetr} \cite{zang2022open} & $0.461\textcolor{darkpink}{\scriptstyle~\diamond44.7\%}$ & $0.307\textcolor{darkpink}{\scriptstyle~\diamond79.1\%}$ & $33.5\textcolor{darkpink}{\scriptstyle~\diamond20.0\%}$ & $0.297\textcolor{darkpink}{\scriptstyle~\diamond156.9\%}$ & $89.1.\textcolor{darkpink}{\scriptstyle~\diamond5.3\%}$ & $66.5\textcolor{darkpink}{\scriptstyle~\diamond36.2\%}$ \\
  \textcolor[rgb]{0.99,0.5,0.0}{OpenSeeD} \cite{zhang2023simple} & $0.525\textcolor{darkpink}{\scriptstyle~\diamond27.0\%}$ & $0.437\textcolor{darkpink}{\scriptstyle~\diamond25.9\%}$ & $36.9\textcolor{darkpink}{\scriptstyle~\diamond8.9\%}$ & $0.487\textcolor{darkpink}{\scriptstyle~\diamond56.7\%}$ & $91.5\textcolor{darkpink}{\scriptstyle~\diamond2.5\%}$ & $84.9\textcolor{darkpink}{\scriptstyle~\diamond6.7\%}$  \\

  \hline 
  \rowcolor{mygray}
  \textbf{Ours}  & $\bm{0.667}_{\pm 0.013}$ & $\bm{0.550}_{\pm 0.03}$ & $\bm{40.2}_{\pm 0.2}$ & $\bm{0.763}_{\pm0.016}$ & $\bm{93.8}_{\pm 0.2}$ & $\bm{90.6}_{\pm 0.1}$ \\
    \hline
    \Xhline{2.\arrayrulewidth}
    \end{tabular}
  \end{table*}

\begin{figure*}[t]
	\centering
		\begin{overpic}[width=1.00\linewidth]{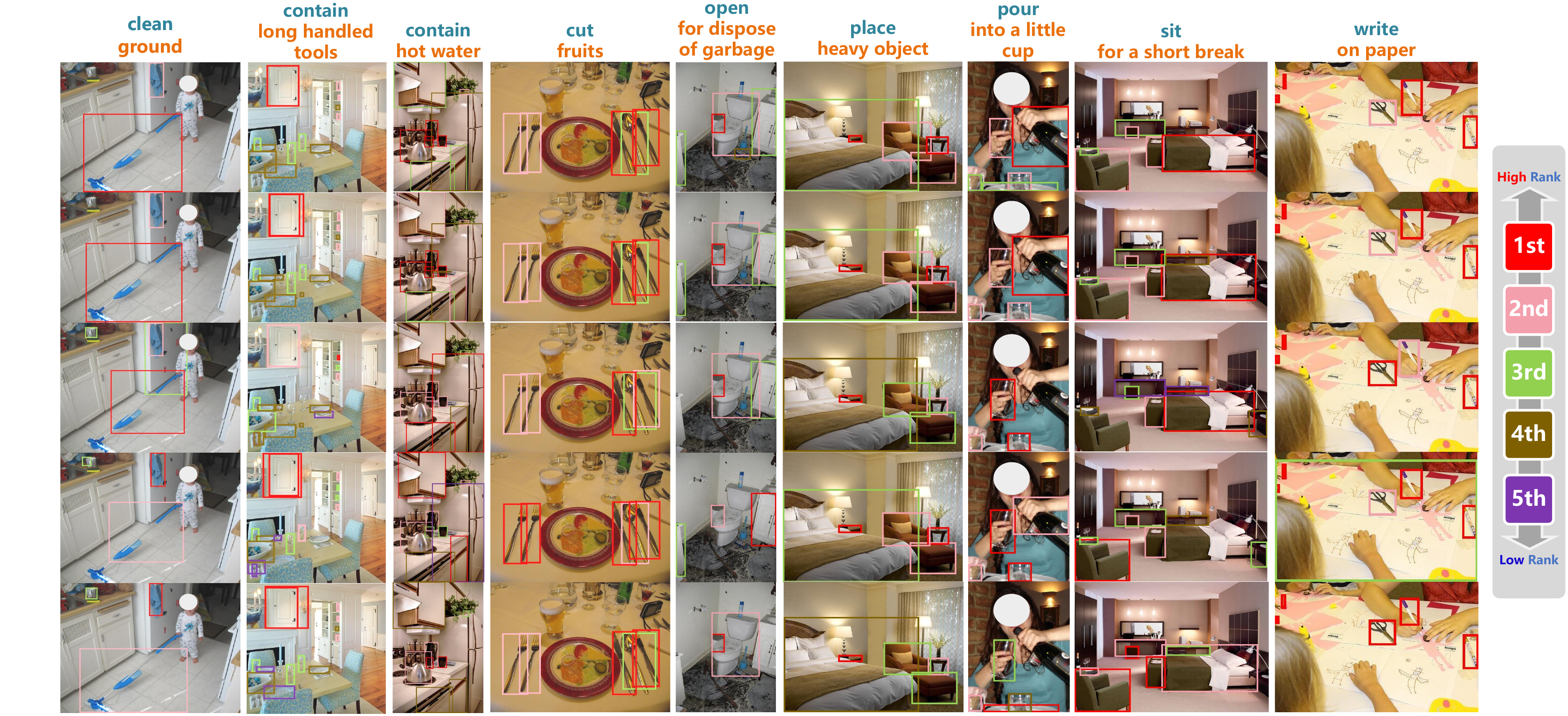}
        \put(2.1,36.4){\rotatebox{90}{\small\textbf{GT}}}
        \put(2.1,27.2){\rotatebox{90}{\small\textbf{Ours}}}
        \put(2.1,18.7){\rotatebox{90}{\small\textbf{\cite{liu2021instance}}}}
        \put(2.1,10.7){\rotatebox{90}{\small\textbf{\cite{deng2024advancing}}}}
         \put(2.1,2.7){\rotatebox{90}{\small\textbf{\cite{zang2022open}}}}
	\end{overpic}
        \captionsetup{aboveskip=0pt}
	\caption{\textbf{Ranking results on the TAR dataset. }We visualize the ranking results for the nine tasks and select IRSR\cite{liu2021instance} and QAGNet\cite{deng2024advancing} from saliency ranking and OVDetr\cite{zang2022open} from multimodal object detection for comparison.
	}
	\label{vis_res}
\end{figure*}

\begin{figure}[t]
	\centering
		\begin{overpic}[width=1.0\linewidth]{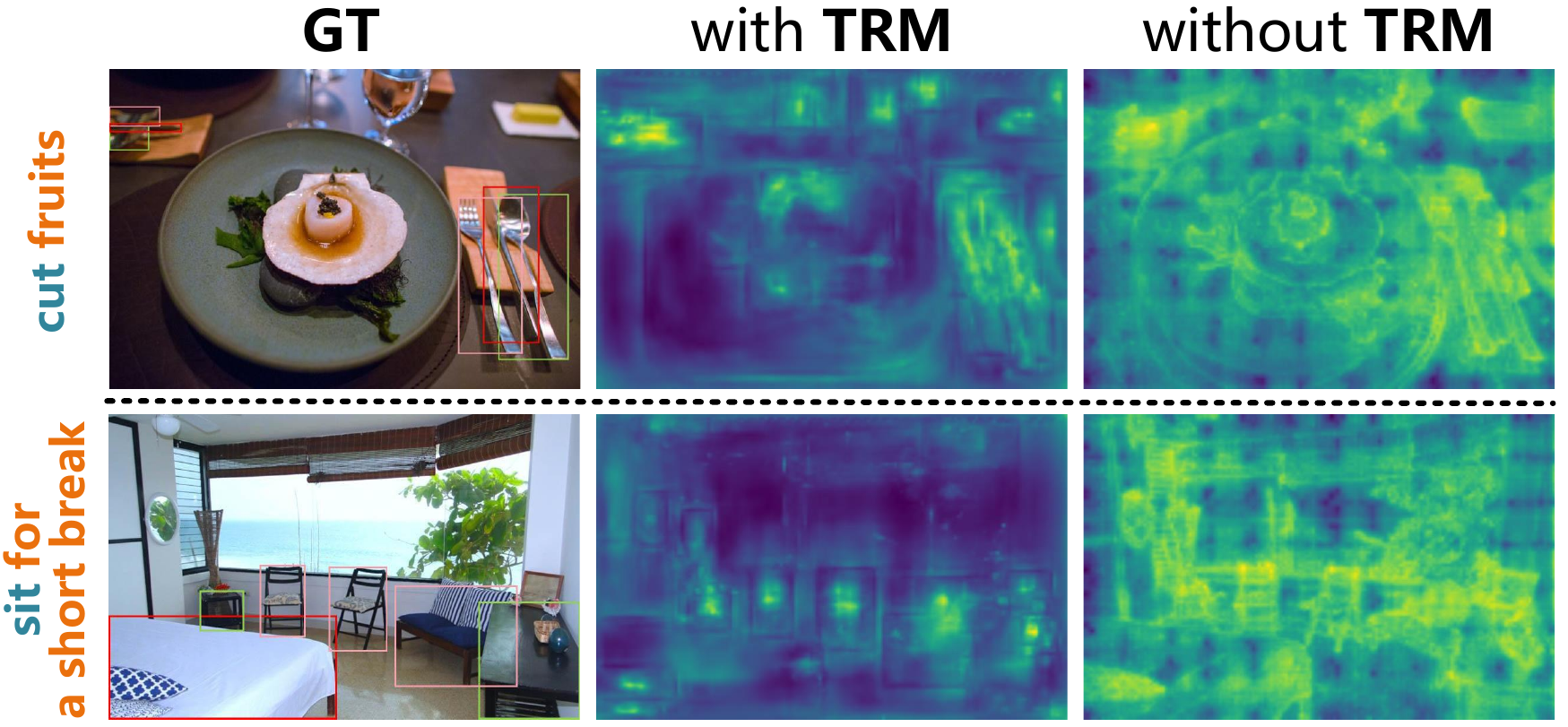}
        
	\end{overpic}
        \captionsetup{aboveskip=4pt}
        \captionsetup{belowskip=0pt}
	\caption{\textbf{Ablation of TRM.} The discriminability score maps\cite{zong2023detrs} with and without the TRM module, calculating from the $l^2$ norm of features at different scales.}
	\label{trm_act}
\end{figure}

\begin{figure}[t]
	\centering
		\begin{overpic}[width=0.93\linewidth]{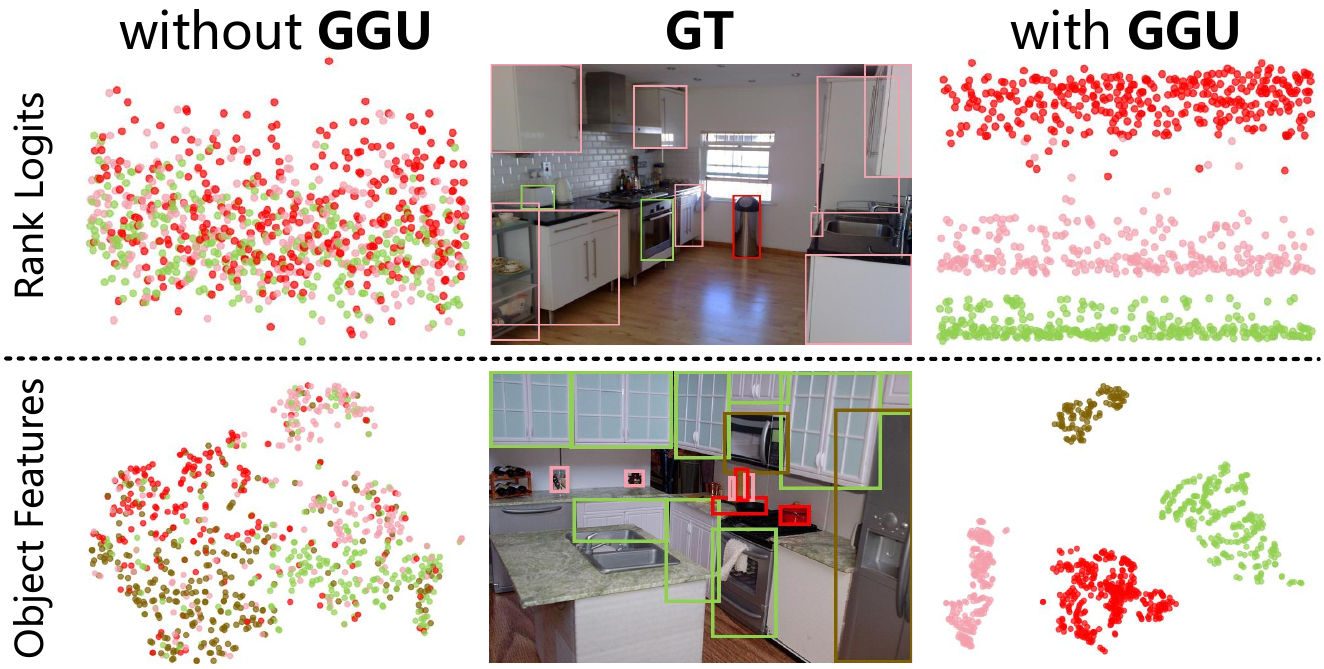}
        
	\end{overpic}
        \captionsetup{aboveskip=2pt}
	\caption{\textbf{Ablation of GGU.} Results with and without the GGU, the top row are ranking logits, with positions higher up indicating higher rank priority, the bottom row are the t-SNE\cite{van2008visualizing} results of object features. }
	\label{vis_ggu}
\end{figure}

\section{Experiments}
\subsection{Benchmark Setting}
We use SSOR \cite{siris2020inferring} and SA-SOR\cite{liu2021instance}, commonly employed in the field of saliency ranking, as evaluation metrics for ranking, mAP@50 as the metric for detection, and the Adjusted Rand Index(ARI) \cite{hubert1985comparing} as the metric for grouping. Additionally, we classify the detected objects into task-relevant and task-irrelevant based on a threshold, and calculate Task-acc and Task-recall to measure the correspondence between the detected objects and the task. Our model is implemented using PyTorch and utilizes the AdamW \cite{loshchilov2017decoupled} optimizer with an initial learning rate of $1e^{-4}$. The number of object queries is set to 900, and the number of group tokens is set to 8, as TAR includes up to 7 ranking levels plus one for irrelevant objects. The hyper-parameters $\lambda_1$, $\lambda_2$, $\lambda_3$ and $\lambda_4$ are set to 2, 2, 5 and 4 respectively. The group penalty parameter $\rho$ is set to 0.5. Our model is trained on 4 A6000 for 25,000 iterations with batch size 8.
Since no previous works have used task descriptions to perform affordance ranking, we select two types of relevant methods for a comprehensive comparison with our approach. The first type is the saliency ranking method, including SOR \cite{fang2021salient}, IRSR \cite{liu2021instance}, and QAGNet \cite{deng2024advancing}. Since these methods do not incorporate text features, we concatenate the extracted text features with the image features and pass them through a simple transformer layer to serve as the baseline model. 
The second type involves multimodal object detection approaches, mdetr \cite{kamath2021mdetr}, OVDetr \cite{zang2022open} and OpenSeeD \cite{zhang2023simple}. Since these methods cannot be directly used for ranking, we add a simple ranking head on top of the object features they extract to output ranking results. 

\subsection{Quantitative and Qualitative Comparisons}
The experimental results and comparisons with other baselines are shown in the Table \ref{ce_res}. It shows that our method achieves the best results across all metrics. For the baseline models in the saliency ranking field, our model achieve improvements of \textbf{34\%}, \textbf{26\%}, and \textbf{31\%} on the SOR metric, and similarly \textbf{71\%}, \textbf{33\%}, and \textbf{36\%} on the SSOR metric, demonstrating its superior ability in modeling rank relationships. As for the advanced baseline models in the multimodal object detection field, our method achieve an average improvement of over \textbf{30\%} on mAP@50, indicating that our model can also accurately extract task-relevant object features. In addition, we also visualize the ranking of detected objects, as shown in the Fig. \ref{vis_res}. For ``clean ground'', our model accurately detect the most suitable broom and the second-best towel, while other models exhibit issues such as poor localization, incorrect selection of irrelevant objects, or faulty ranking. Besides, for examples within complex scenes, our model also shows better detection and ranking performance. In contrast, IRSR\cite{liu2021instance} and OVDetr\cite{zang2022open} tend to incorrectly group objects or miss detecting some objects. The results prove that through deep task context integration and group-wise global updates, our model can efficiently detects task-related objects and exactly rank their priority. 


\begin{table}[!t]
\centering
  \footnotesize
  \renewcommand{\arraystretch}{1.}
  \renewcommand{\tabcolsep}{4.0 pt}
  \captionsetup{aboveskip=1pt}
  \caption{\textbf{Ablation study of different modules.} We investigate the effectiveness of TRM module and GGU strategy on our TAR.}
   \label{ablation_model}
   \begin{tabular}{ccc||ccc}
       \hline
    \Xhline{2.\arrayrulewidth}
    \textbf{TRM} & $\textbf{GGU(Group)}$ & $\textbf{GGU(Graph)}$ & $\textbf{SSOR}$ & $\textbf{mAP@50}$ & \textbf{ARI} \\
      \Xhline{2.\arrayrulewidth}
    \multirow{8}{*}  & & & $0.312$  & $22.9$ & $0.310$ \\
   $\checkmark$ & & & $0.451$ & $36.7$ & $0.564$\\
     & $\checkmark$ &   & $0.375$ & $29.7$ & $0.377$ \\
    &  & $\checkmark$ & $0.430$ & $30.0$ & $0.464$ \\
  
   $\checkmark$ & $\checkmark$ &  & $0.492$ & $39.8$ & $0.733$ \\
   $\checkmark$ &  & $\checkmark$ & $0.530$ & $38.5$ & $0.582$ \\
   & $\checkmark$ & $\checkmark$ & $0.521$ & $32.5$ & $0.697$ \\
   \rowcolor{mygray}
    $\checkmark$ & $\checkmark$ & $\checkmark$  & \bm{$0.550$} & \bm{$40.2$} & \bm{$0.763$} \\
    \hline
    \Xhline{2.\arrayrulewidth}
   \end{tabular}

\end{table}

\begin{figure}[t]
	\centering
		\begin{overpic}[width=0.93\linewidth]{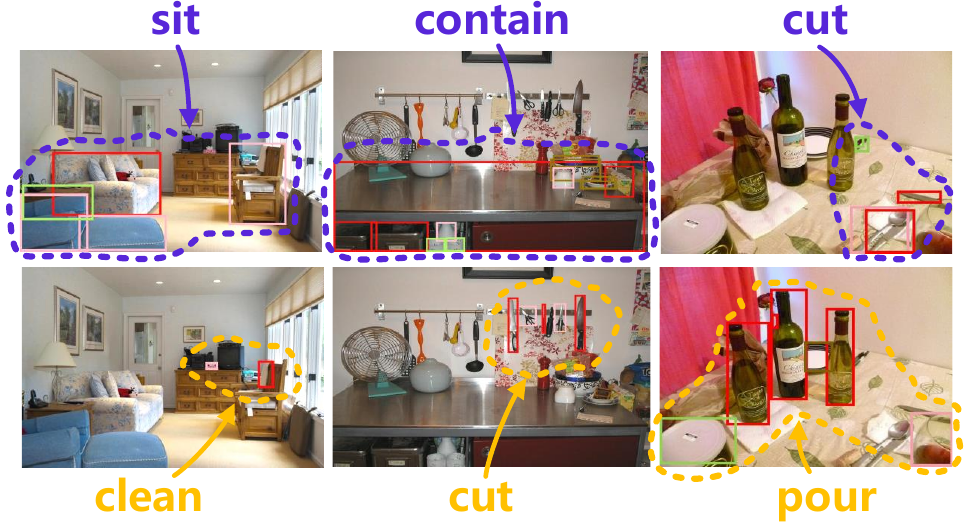}
        
	\end{overpic}
        \captionsetup{aboveskip=2pt}
        \captionsetup{belowskip=0pt}
	\caption{\textbf{Different affordance.} For different affordance, the model will focus on different objects. }
	\label{diff_aff}
\end{figure}

\subsection{Ablation Study}
\begin{figure}[t]
	\centering
		\begin{overpic}[width=0.88\linewidth]{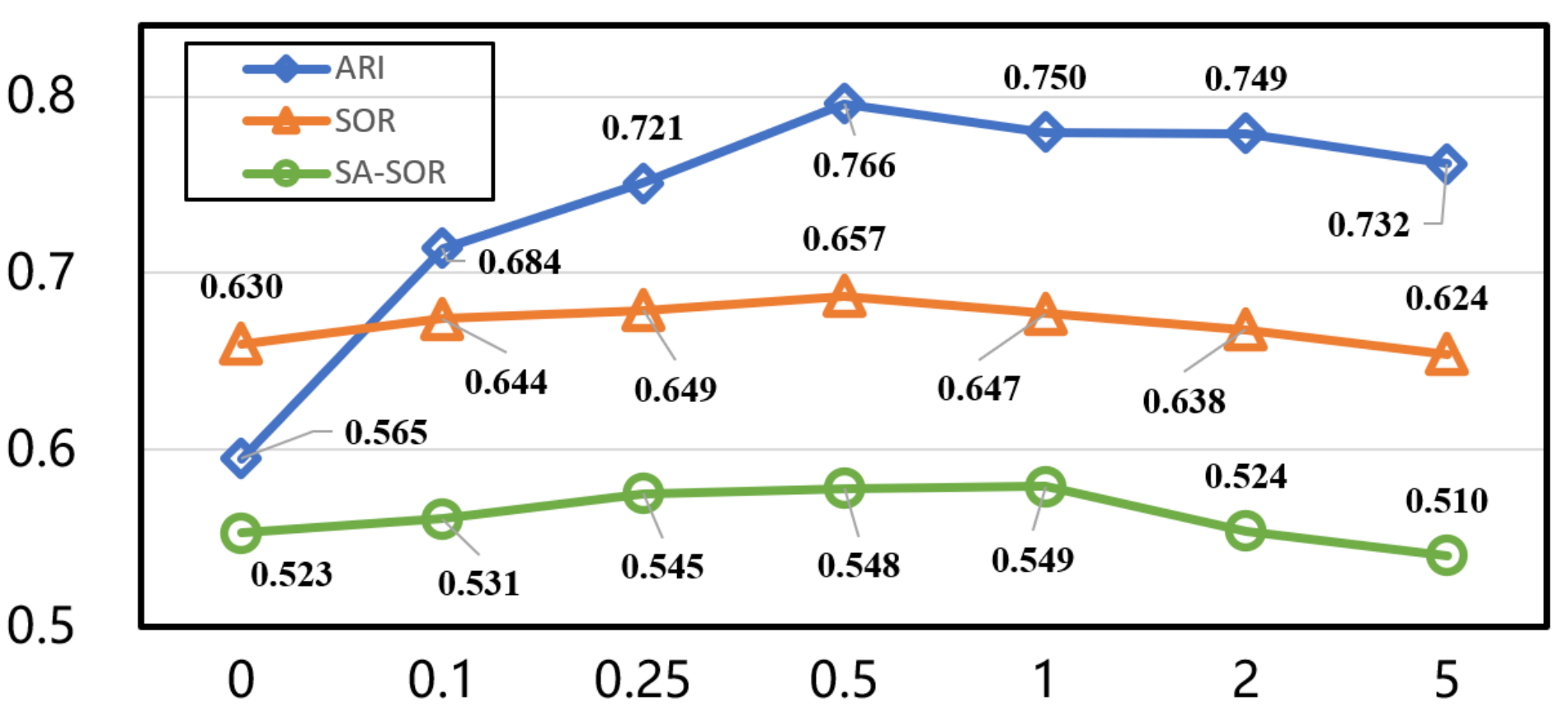}
        
	\end{overpic}
        \captionsetup{aboveskip=0pt}
	\caption{\textbf{Ablation of proposed loss.} We vary the parameter $\rho$ to observe the effectiveness of proposed GRL. }
	\label{ablation_GRL}
\end{figure}

The impact of our proposed modules on the evaluation metrics is shown in Tab \ref{ablation_model}. To validate the effectiveness of our method at a deeper level, beyond that, we detailedly conduct ablation study from two aspects. We first compare the discriminability score map\cite{zong2023detrs} of latent features with and without the TRM module to highlight TRM's ability to effectively utilize task cues in guiding the detection of relevant objects, as shown in Fig. \ref{trm_act}. It shows that with the TRM module, the model's focus is primarily directed toward task-related objects, which not only reduces unnecessary computations but also lays a solid foundation for subsequent ranking calculations. Then, we visualize the rank logits of all candidate objects to demonstrate the grouping and ranking capability of the GGU module, as shown in Fig. \ref{vis_ggu}. It indicates that with the GGU module, the ranking logits of all candidate objects exhibit a clear hierarchical order of priority levels. We also visualize the features of object queries using t-SNE\cite{van2008visualizing} in Fig. \ref{vis_ggu}, which further demonstrate that GGU can accurately identify functional groups within the scene to complete our ranking task. Besides, We conduct the ablation study of our proposed group ranking loss in Fig. \ref{ablation_GRL}. It shows that compared to setting $\rho$=0\cite{liu2021instance}, the ARI metric increases significantly as the $\rho$ value rises, and stabilizes after $\rho$ reaches 0.5. Other metrics also show improvement, which prove that our proposed GRL can facilitate the training of the group ranking model.



\subsection{Performance Analysis}

\textbf{Objects Change With Affordance.\ }Different affordances correspond to various selectable objects, and our dataset contains many instances where a single image corresponds to multiple affordances.  Fig. \ref{diff_aff} indicates that our model can correctly attend to different objects based on varying affordance, which aids robots in making swift and right initial judgments in complex environments. 

\begin{figure}[t]
	\centering
		\begin{overpic}[width=0.93\linewidth]{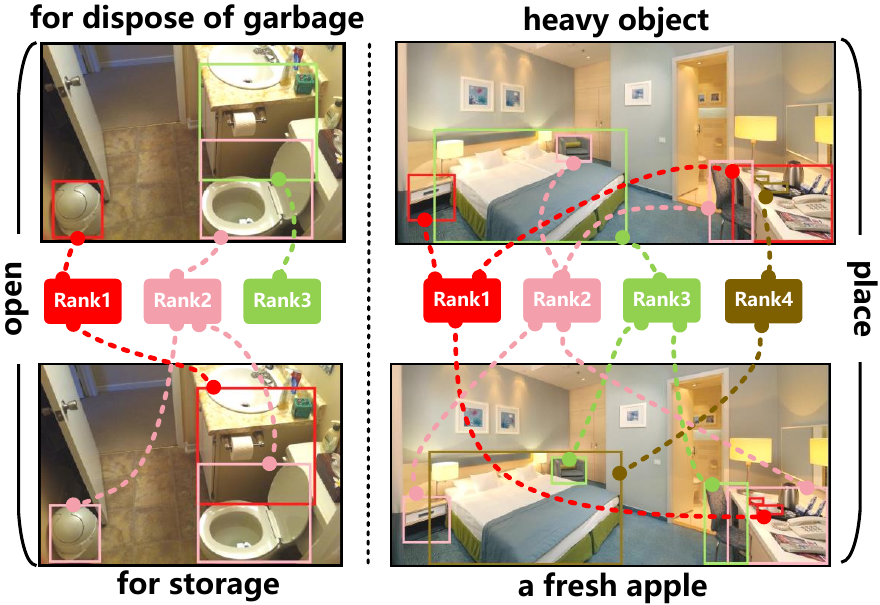}
        
	\end{overpic}
        \captionsetup{aboveskip=0pt}
	\caption{\textbf{Different context.} For the same affordance under different task contexts, both the predicted ranks and the functional groups vary. }
	\label{diff_con}
\end{figure}

\textbf{Ranks Change With Task Context.\ }For an intelligent agent, the ability to recognize objects from different perspectives based on varying task contexts is crucial. Fig. \ref{diff_con} illustrates that our model can appropriately group objects and provide corresponding rankings under different contexts, highlighting its practical utility.

\begin{figure}[t]
	\centering
		\begin{overpic}[width=0.95\linewidth]{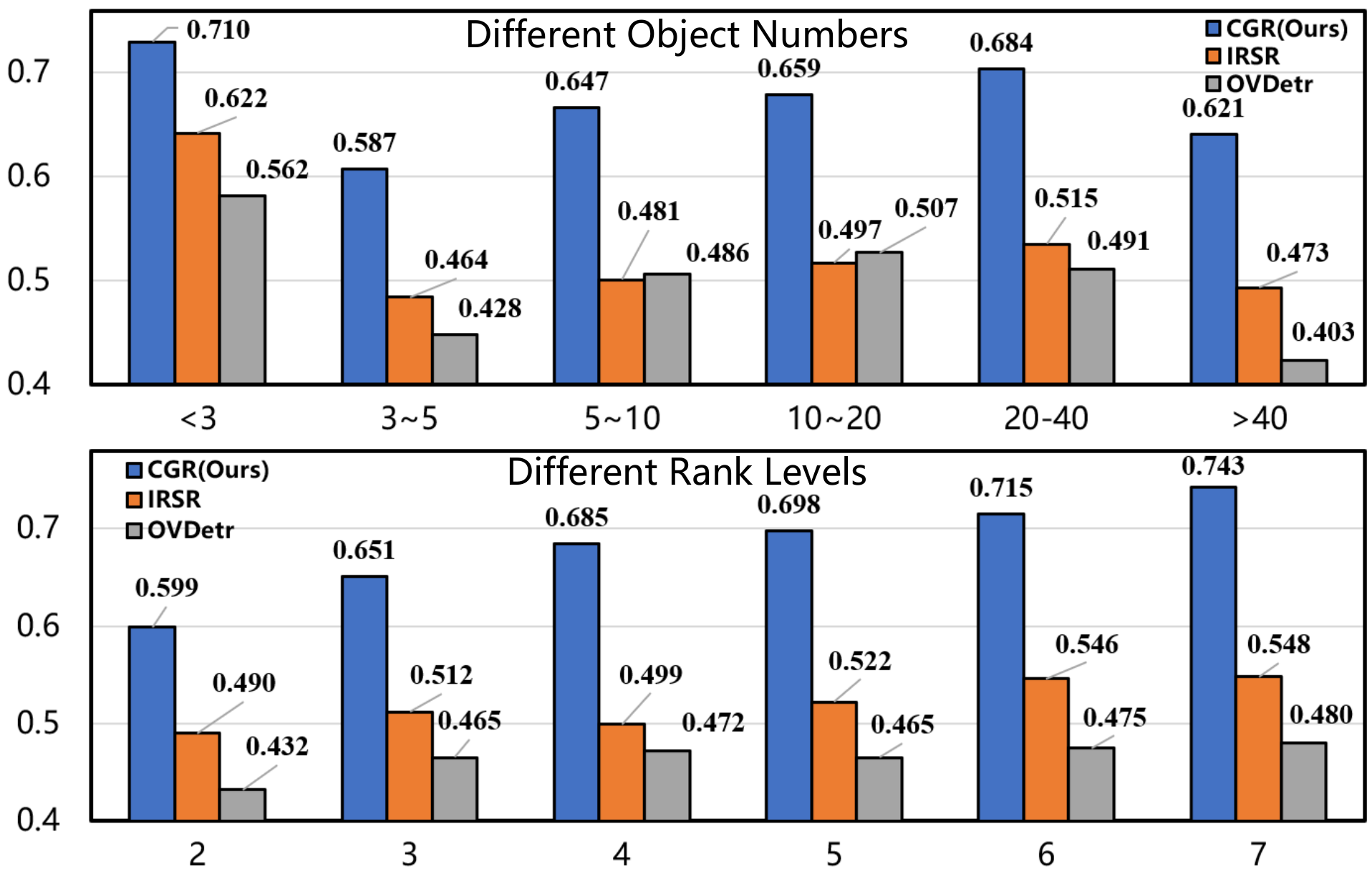}
        
	\end{overpic}
        \captionsetup{aboveskip=2pt}
	\caption{\textbf{Different object numbers and rank levels.} We calculate and compare the SOR metrics under instances with different object numbers and ranking levels.}
	\label{diff_obj_level}
\end{figure}

\textbf{Different Object Numbers and Rank Levels.\ }
We calculate the SOR of instances with different object numbers and ranking levels, and compare with other methods in Fig. \ref{diff_obj_level}.
As the number of objects increases, the SOR exhibits an upward trend, which becomes more pronounced with the ranking level increases. Regarding this phenomenon, we infer that a larger number of objects provides richer semantic information to guide the ranking, which also indirectly demonstrates the reliability of our dataset.

\textbf{Limitations.\ }
In our setting, we first detect all objects and then use relational operations followed by a threshold to filter out task-relevant objects. Although our model demonstrates decent task accuracy, this method may introduce some noise compared to directly detecting task-relevant objects, which can affect the final performance. In the future, we will enhance the method to directly achieve the ranking of task-relevant objects in a single step. 

\section{Conclusion}
In this paper, we explore a new paradigm for affordance learning: leverage task context for object affordance ranking, aiming at enabling agents to select appropriate objects based on task requirements.
Furthermore, we collect the TAR dataset to support this task, which includes bounding box annotations for task-relevant objects along with their corresponding ranking annotations. We also present a new group ranking model to tackle this challenging task through deep task integration and global information transmission within groups, achieving state-of-the-art performance on the proposed TAR dataset. 

{\small
\bibliographystyle{ieee_fullname}
\bibliography{egbib}

\begin{thebibliography}{10}\itemsep=-1pt

\bibitem{ahn2022can}
Michael Ahn, Anthony Brohan, Noah Brown, Yevgen Chebotar, Omar Cortes, Byron David, Chelsea Finn, Chuyuan Fu, Keerthana Gopalakrishnan, Karol Hausman, et~al.
\newblock Do as i can, not as i say: Grounding language in robotic affordances.
\newblock {\em arXiv preprint arXiv:2204.01691}, 2022.

\bibitem{bahl2023affordances}
Shikhar Bahl, Russell Mendonca, Lili Chen, Unnat Jain, and Deepak Pathak.
\newblock Affordances from human videos as a versatile representation for robotics.
\newblock In {\em Proceedings of the IEEE/CVF Conference on Computer Vision and Pattern Recognition}, pages 13778--13790, 2023.

\bibitem{carion2020end}
Nicolas Carion, Francisco Massa, Gabriel Synnaeve, Nicolas Usunier, Alexander Kirillov, and Sergey Zagoruyko.
\newblock End-to-end object detection with transformers.
\newblock In {\em European conference on computer vision}, pages 213--229. Springer, 2020.

\bibitem{chen2024taskclip}
Hanning Chen, Wenjun Huang, Yang Ni, Sanggeon Yun, Fei Wen, Hugo Latapie, and Mohsen Imani.
\newblock Taskclip: Extend large vision-language model for task oriented object detection.
\newblock {\em arXiv preprint arXiv:2403.08108}, 2024.

\bibitem{chen2024vltp}
Hanning Chen, Yang Ni, Wenjun Huang, Yezi Liu, SungHeon Jeong, Fei Wen, Nathaniel Bastian, Hugo Latapie, and Mohsen Imani.
\newblock Vltp: Vision-language guided token pruning for task-oriented segmentation.
\newblock {\em arXiv preprint arXiv:2409.08464}, 2024.

\bibitem{chen2023affordance}
Joya Chen, Difei Gao, Kevin~Qinghong Lin, and Mike~Zheng Shou.
\newblock Affordance grounding from demonstration video to target image.
\newblock In {\em Proceedings of the IEEE/CVF Conference on Computer Vision and Pattern Recognition}, pages 6799--6808, 2023.

\bibitem{chuang2018learning}
Ching-Yao Chuang, Jiaman Li, Antonio Torralba, and Sanja Fidler.
\newblock Learning to act properly: Predicting and explaining affordances from images.
\newblock In {\em Proceedings of the IEEE Conference on Computer Vision and Pattern Recognition}, pages 975--983, 2018.

\bibitem{deng2024advancing}
Bowen Deng, Siyang Song, Andrew~P French, Denis Schluppeck, and Michael~P Pound.
\newblock Advancing saliency ranking with human fixations: Dataset models and benchmarks.
\newblock In {\em Proceedings of the IEEE/CVF Conference on Computer Vision and Pattern Recognition}, pages 28348--28357, 2024.

\bibitem{devlin2018bert}
Jacob Devlin.
\newblock Bert: Pre-training of deep bidirectional transformers for language understanding.
\newblock {\em arXiv preprint arXiv:1810.04805}, 2018.

\bibitem{do2018affordancenet}
Thanh-Toan Do, Anh Nguyen, and Ian Reid.
\newblock Affordancenet: An end-to-end deep learning approach for object affordance detection.
\newblock In {\em 2018 IEEE international conference on robotics and automation (ICRA)}, pages 5882--5889. IEEE, 2018.

\bibitem{fang2021salient}
Hao Fang, Daoxin Zhang, Yi Zhang, Minghao Chen, Jiawei Li, Yao Hu, Deng Cai, and Xiaofei He.
\newblock Salient object ranking with position-preserved attention.
\newblock In {\em Proceedings of the IEEE/CVF International Conference on Computer Vision}, pages 16331--16341, 2021.

\bibitem{ge2024behavior}
Yunhao Ge, Yihe Tang, Jiashu Xu, Cem Gokmen, Chengshu Li, Wensi Ai, Benjamin~Jose Martinez, Arman Aydin, Mona Anvari, Ayush~K Chakravarthy, et~al.
\newblock Behavior vision suite: Customizable dataset generation via simulation.
\newblock In {\em Proceedings of the IEEE/CVF Conference on Computer Vision and Pattern Recognition}, pages 22401--22412, 2024.

\bibitem{huang2018holistic}
Siyuan Huang, Siyuan Qi, Yixin Zhu, Yinxue Xiao, Yuanlu Xu, and Song-Chun Zhu.
\newblock Holistic 3d scene parsing and reconstruction from a single rgb image.
\newblock In {\em Proceedings of the European conference on computer vision (ECCV)}, pages 187--203, 2018.

\bibitem{hubert1985comparing}
Lawrence Hubert and Phipps Arabie.
\newblock Comparing partitions.
\newblock {\em Journal of classification}, 2:193--218, 1985.

\bibitem{islam2018revisiting}
Md~Amirul Islam, Mahmoud Kalash, and Neil~DB Bruce.
\newblock Revisiting salient object detection: Simultaneous detection, ranking, and subitizing of multiple salient objects.
\newblock In {\em Proceedings of the IEEE conference on computer vision and pattern recognition}, pages 7142--7150, 2018.

\bibitem{jang2016categorical}
Eric Jang, Shixiang Gu, and Ben Poole.
\newblock Categorical reparameterization with gumbel-softmax.
\newblock {\em arXiv preprint arXiv:1611.01144}, 2016.

\bibitem{kahneman1973attention}
Daniel Kahneman.
\newblock Attention and effort, 1973.

\bibitem{kamath2021mdetr}
Aishwarya Kamath, Mannat Singh, Yann LeCun, Gabriel Synnaeve, Ishan Misra, and Nicolas Carion.
\newblock Mdetr-modulated detection for end-to-end multi-modal understanding.
\newblock In {\em Proceedings of the IEEE/CVF international conference on computer vision}, pages 1780--1790, 2021.

\bibitem{koppula2014physically}
Hema~S Koppula and Ashutosh Saxena.
\newblock Physically grounded spatio-temporal object affordances.
\newblock In {\em Computer Vision--ECCV 2014: 13th European Conference, Zurich, Switzerland, September 6-12, 2014, Proceedings, Part III 13}, pages 831--847. Springer, 2014.

\bibitem{lamdouar2023making}
Hala Lamdouar, Weidi Xie, and Andrew Zisserman.
\newblock The making and breaking of camouflage.
\newblock In {\em Proceedings of the IEEE/CVF international conference on computer vision}, pages 832--842, 2023.

\bibitem{li2023locate}
Gen Li, Varun Jampani, Deqing Sun, and Laura Sevilla-Lara.
\newblock Locate: Localize and transfer object parts for weakly supervised affordance grounding.
\newblock In {\em Proceedings of the IEEE/CVF Conference on Computer Vision and Pattern Recognition}, pages 10922--10931, 2023.

\bibitem{li2022toist}
Pengfei Li, Beiwen Tian, Yongliang Shi, Xiaoxue Chen, Hao Zhao, Guyue Zhou, and Ya-Qin Zhang.
\newblock Toist: Task oriented instance segmentation transformer with noun-pronoun distillation.
\newblock {\em Advances in Neural Information Processing Systems}, 35:17597--17611, 2022.

\bibitem{lin2022rethinking}
Jiaying Lin, Huankang Guan, and Rynson~WH Lau.
\newblock Rethinking video salient object ranking.
\newblock {\em arXiv preprint arXiv:2203.17257}, 2022.

\bibitem{liu2021instance}
Nian Liu, Long Li, Wangbo Zhao, Junwei Han, and Ling Shao.
\newblock Instance-level relative saliency ranking with graph reasoning.
\newblock {\em IEEE Transactions on Pattern Analysis and Machine Intelligence}, 44(11):8321--8337, 2021.

\bibitem{liu2022joint}
Shaowei Liu, Subarna Tripathi, Somdeb Majumdar, and Xiaolong Wang.
\newblock Joint hand motion and interaction hotspots prediction from egocentric videos.
\newblock In {\em Proceedings of the IEEE/CVF Conference on Computer Vision and Pattern Recognition}, pages 3282--3292, 2022.

\bibitem{liu2023grounding}
Shilong Liu, Zhaoyang Zeng, Tianhe Ren, Feng Li, Hao Zhang, Jie Yang, Qing Jiang, Chunyuan Li, Jianwei Yang, Hang Su, et~al.
\newblock Grounding dino: Marrying dino with grounded pre-training for open-set object detection.
\newblock {\em arXiv preprint arXiv:2303.05499}, 2023.

\bibitem{liu2017single}
Xiaobai Liu, Yibiao Zhao, and Song-Chun Zhu.
\newblock Single-view 3d scene reconstruction and parsing by attribute grammar.
\newblock {\em IEEE transactions on pattern analysis and machine intelligence}, 40(3):710--725, 2017.

\bibitem{liu2021swin}
Ze Liu, Yutong Lin, Yue Cao, Han Hu, Yixuan Wei, Zheng Zhang, Stephen Lin, and Baining Guo.
\newblock Swin transformer: Hierarchical vision transformer using shifted windows.
\newblock In {\em Proceedings of the IEEE/CVF international conference on computer vision}, pages 10012--10022, 2021.

\bibitem{loshchilov2017decoupled}
I Loshchilov.
\newblock Decoupled weight decay regularization.
\newblock {\em arXiv preprint arXiv:1711.05101}, 2017.

\bibitem{lu2022phrase}
Liangsheng Lu, Wei Zhai, Hongchen Luo, Yu Kang, and Yang Cao.
\newblock Phrase-based affordance detection via cyclic bilateral interaction.
\newblock {\em IEEE Transactions on Artificial Intelligence}, 4(5):1186--1198, 2022.

\bibitem{luo2021one}
Hongchen Luo, Wei Zhai, Jing Zhang, Yang Cao, and Dacheng Tao.
\newblock One-shot affordance detection.
\newblock {\em arXiv preprint arXiv:2106.14747}, 2021.

\bibitem{luo2022learning}
Hongchen Luo, Wei Zhai, Jing Zhang, Yang Cao, and Dacheng Tao.
\newblock Learning affordance grounding from exocentric images.
\newblock In {\em Proceedings of the IEEE/CVF conference on computer vision and pattern recognition}, pages 2252--2261, 2022.

\bibitem{luo2023leverage}
Hongchen Luo, Wei Zhai, Jing Zhang, Yang Cao, and Dacheng Tao.
\newblock Leverage interactive affinity for affordance learning.
\newblock In {\em Proceedings of the IEEE/CVF Conference on Computer Vision and Pattern Recognition}, pages 6809--6819, 2023.

\bibitem{lv2021simultaneously}
Yunqiu Lv, Jing Zhang, Yuchao Dai, Aixuan Li, Bowen Liu, Nick Barnes, and Deng-Ping Fan.
\newblock Simultaneously localize, segment and rank the camouflaged objects.
\newblock In {\em Proceedings of the IEEE/CVF conference on computer vision and pattern recognition}, pages 11591--11601, 2021.

\bibitem{myers2015affordance}
Austin Myers, Ching~L Teo, Cornelia Ferm{\"u}ller, and Yiannis Aloimonos.
\newblock Affordance detection of tool parts from geometric features.
\newblock In {\em 2015 IEEE International Conference on Robotics and Automation (ICRA)}, pages 1374--1381. IEEE, 2015.

\bibitem{nagarajan2019grounded}
Tushar Nagarajan, Christoph Feichtenhofer, and Kristen Grauman.
\newblock Grounded human-object interaction hotspots from video.
\newblock In {\em Proceedings of the IEEE/CVF International Conference on Computer Vision}, pages 8688--8697, 2019.

\bibitem{nagarajan2020learning}
Tushar Nagarajan and Kristen Grauman.
\newblock Learning affordance landscapes for interaction exploration in 3d environments.
\newblock {\em Advances in Neural Information Processing Systems}, 33:2005--2015, 2020.

\bibitem{nguyen2017object}
Anh Nguyen, Dimitrios Kanoulas, Darwin~G Caldwell, and Nikos~G Tsagarakis.
\newblock Object-based affordances detection with convolutional neural networks and dense conditional random fields.
\newblock In {\em 2017 IEEE/RSJ International Conference on Intelligent Robots and Systems (IROS)}, pages 5908--5915. IEEE, 2017.

\bibitem{pieropan2015functional}
Alessandro Pieropan, Carl~Henrik Ek, and Hedvig Kjellstr{\"o}m.
\newblock Functional descriptors for object affordances.
\newblock In {\em IROS 2015 Workshop}, 2015.

\bibitem{qiao2024hypersor}
Minglang Qiao, Mai Xu, Lai Jiang, Peng Lei, Shijie Wen, Yunjin Chen, and Leonid Sigal.
\newblock Hypersor: Context-aware graph hypernetwork for salient object ranking.
\newblock {\em IEEE Transactions on Pattern Analysis and Machine Intelligence}, 2024.

\bibitem{qu2024rio}
Mengxue Qu, Yu Wu, Wu Liu, Xiaodan Liang, Jingkuan Song, Yao Zhao, and Yunchao Wei.
\newblock Rio: A benchmark for reasoning intention-oriented objects in open environments.
\newblock {\em Advances in Neural Information Processing Systems}, 36, 2024.

\bibitem{ren2024grounding}
Tianhe Ren, Qing Jiang, Shilong Liu, Zhaoyang Zeng, Wenlong Liu, Han Gao, Hongjie Huang, Zhengyu Ma, Xiaoke Jiang, Yihao Chen, et~al.
\newblock Grounding dino 1.5: Advance the" edge" of open-set object detection.
\newblock {\em arXiv preprint arXiv:2405.10300}, 2024.

\bibitem{rezatofighi2019generalized}
Hamid Rezatofighi, Nathan Tsoi, JunYoung Gwak, Amir Sadeghian, Ian Reid, and Silvio Savarese.
\newblock Generalized intersection over union: A metric and a loss for bounding box regression.
\newblock In {\em Proceedings of the IEEE/CVF conference on computer vision and pattern recognition}, pages 658--666, 2019.

\bibitem{ross2017focal}
T-YLPG Ross and GKHP Doll{\'a}r.
\newblock Focal loss for dense object detection.
\newblock In {\em proceedings of the IEEE conference on computer vision and pattern recognition}, pages 2980--2988, 2017.

\bibitem{roy2016multi}
Anirban Roy and Sinisa Todorovic.
\newblock A multi-scale cnn for affordance segmentation in rgb images.
\newblock In {\em Computer Vision--ECCV 2016: 14th European Conference, Amsterdam, The Netherlands, October 11--14, 2016, Proceedings, Part IV 14}, pages 186--201. Springer, 2016.

\bibitem{sawatzky2017adaptive}
Johann Sawatzky and Jurgen Gall.
\newblock Adaptive binarization for weakly supervised affordance segmentation.
\newblock In {\em Proceedings of the IEEE international conference on computer vision workshops}, pages 1383--1391, 2017.

\bibitem{sawatzky2019object}
Johann Sawatzky, Yaser Souri, Christian Grund, and Jurgen Gall.
\newblock What object should i use?-task driven object detection.
\newblock In {\em Proceedings of the IEEE/CVF Conference on Computer Vision and Pattern Recognition}, pages 7605--7614, 2019.

\bibitem{shao2019objects365}
Shuai Shao, Zeming Li, Tianyuan Zhang, Chao Peng, Gang Yu, Xiangyu Zhang, Jing Li, and Jian Sun.
\newblock Objects365: A large-scale, high-quality dataset for object detection.
\newblock In {\em Proceedings of the IEEE/CVF international conference on computer vision}, pages 8430--8439, 2019.

\bibitem{shin2022does}
Donghee Shin.
\newblock Does augmented reality augment user affordance? the effect of technological characteristics on game behaviour.
\newblock {\em Behaviour \& Information Technology}, 41(11):2373--2389, 2022.

\bibitem{siris2020inferring}
Avishek Siris, Jianbo Jiao, Gary~KL Tam, Xianghua Xie, and Rynson~WH Lau.
\newblock Inferring attention shift ranks of objects for image saliency.
\newblock In {\em Proceedings of the IEEE/CVF conference on computer vision and pattern recognition}, pages 12133--12143, 2020.

\bibitem{steffen2019framework}
Jacob~H Steffen, James~E Gaskin, Thomas~O Meservy, Jeffrey~L Jenkins, and Iopa Wolman.
\newblock Framework of affordances for virtual reality and augmented reality.
\newblock {\em Journal of management information systems}, 36(3):683--729, 2019.

\bibitem{sun2010learning}
Jie Sun, Joshua~L Moore, Aaron Bobick, and James~M Rehg.
\newblock Learning visual object categories for robot affordance prediction.
\newblock {\em The International Journal of Robotics Research}, 29(2-3):174--197, 2010.

\bibitem{tang2023cotdet}
Jiajin Tang, Ge Zheng, Jingyi Yu, and Sibei Yang.
\newblock Cotdet: Affordance knowledge prompting for task driven object detection.
\newblock In {\em Proceedings of the IEEE/CVF International Conference on Computer Vision}, pages 3068--3078, 2023.

\bibitem{ugur2011going}
Emre Ugur, Erhan Oztop, and Erol {\c{S}}ahin.
\newblock Going beyond the perception of affordances: Learning how to actualize them through behavioral parameters.
\newblock In {\em 2011 IEEE International Conference on Robotics and Automation}, pages 4768--4773. IEEE, 2011.

\bibitem{van2008visualizing}
Laurens Van~der Maaten and Geoffrey Hinton.
\newblock Visualizing data using t-sne.
\newblock {\em Journal of machine learning research}, 9(11), 2008.

\bibitem{varadarajan2012afrob}
Karthik~Mahesh Varadarajan and Markus Vincze.
\newblock Afrob: The affordance network ontology for robots.
\newblock In {\em 2012 IEEE/RSJ international conference on intelligent robots and systems}, pages 1343--1350. IEEE, 2012.

\bibitem{wolfel2018grounding}
Kim W{\"o}lfel and Dominik Henrich.
\newblock Grounding verbs for tool-dependent, sensor-based robot tasks.
\newblock In {\em 2018 27th IEEE international symposium on robot and human interactive communication (RO-MAN)}, pages 378--383. IEEE, 2018.

\bibitem{xu2022groupvit}
Jiarui Xu, Shalini De~Mello, Sifei Liu, Wonmin Byeon, Thomas Breuel, Jan Kautz, and Xiaolong Wang.
\newblock Groupvit: Semantic segmentation emerges from text supervision.
\newblock In {\em Proceedings of the IEEE/CVF Conference on Computer Vision and Pattern Recognition}, pages 18134--18144, 2022.

\bibitem{yang2023recent}
Xintong Yang, Ze Ji, Jing Wu, and Yu-Kun Lai.
\newblock Recent advances of deep robotic affordance learning: a reinforcement learning perspective.
\newblock {\em IEEE Transactions on Cognitive and Developmental Systems}, 15(3):1139--1149, 2023.

\bibitem{yang2024egochoir}
Yuhang Yang, Wei Zhai, Chengfeng Wang, Chengjun Yu, Yang Cao, and Zheng-Jun Zha.
\newblock Egochoir: Capturing 3d human-object interaction regions from egocentric views.
\newblock {\em arXiv preprint arXiv:2405.13659}, 2024.

\bibitem{ye2017can}
Chengxi Ye, Yezhou Yang, Ren Mao, Cornelia Ferm{\"u}ller, and Yiannis Aloimonos.
\newblock What can i do around here? deep functional scene understanding for cognitive robots.
\newblock In {\em 2017 IEEE International Conference on Robotics and Automation (ICRA)}, pages 4604--4611. IEEE, 2017.

\bibitem{yildirim2020evaluating}
G{\"o}khan Yildirim, Debashis Sen, Mohan Kankanhalli, and Sabine S{\"u}sstrunk.
\newblock Evaluating salient object detection in natural images with multiple objects having multi-level saliency.
\newblock {\em IET Image Processing}, 14(10):2249--2262, 2020.

\bibitem{zang2022open}
Yuhang Zang, Wei Li, Kaiyang Zhou, Chen Huang, and Chen~Change Loy.
\newblock Open-vocabulary detr with conditional matching.
\newblock In {\em European Conference on Computer Vision}, pages 106--122. Springer, 2022.

\bibitem{zhai2022one}
Wei Zhai, Hongchen Luo, Jing Zhang, Yang Cao, and Dacheng Tao.
\newblock One-shot object affordance detection in the wild.
\newblock {\em International Journal of Computer Vision}, 130(10):2472--2500, 2022.

\bibitem{zhang2023simple}
Hao Zhang, Feng Li, Xueyan Zou, Shilong Liu, Chunyuan Li, Jianwei Yang, and Lei Zhang.
\newblock A simple framework for open-vocabulary segmentation and detection.
\newblock In {\em Proceedings of the IEEE/CVF International Conference on Computer Vision}, pages 1020--1031, 2023.

\bibitem{zhao2013scene}
Yibiao Zhao and Song-Chun Zhu.
\newblock Scene parsing by integrating function, geometry and appearance models.
\newblock In {\em Proceedings of the IEEE conference on computer vision and pattern recognition}, pages 3119--3126, 2013.

\bibitem{zheng2018affordances}
Ling Zheng, Tao Xie, and Geping Liu.
\newblock Affordances of virtual reality for collaborative learning.
\newblock In {\em 2018 international joint conference on information, media and engineering (ICIME)}, pages 6--10. IEEE, 2018.

\bibitem{zhu2015understanding}
Yixin Zhu, Yibiao Zhao, and Song Chun~Zhu.
\newblock Understanding tools: Task-oriented object modeling, learning and recognition.
\newblock In {\em Proceedings of the IEEE Conference on Computer Vision and Pattern Recognition}, pages 2855--2864, 2015.

\bibitem{zong2023detrs}
Zhuofan Zong, Guanglu Song, and Yu Liu.
\newblock Detrs with collaborative hybrid assignments training.
\newblock In {\em Proceedings of the IEEE/CVF international conference on computer vision}, pages 6748--6758, 2023.

\end{thebibliography}
}

\end{document}